\documentclass[10pt,journal,compsoc]{IEEEtran}

\ifCLASSOPTIONcompsoc
  \usepackage[nocompress]{cite}
\else
  \usepackage{cite}
\fi

\usepackage[square, comma, sort&compress, numbers]{natbib}

\usepackage{tabu}
\usepackage{threeparttable}
\usepackage{multirow}
\usepackage{booktabs}
\usepackage{amsmath}
\usepackage{amsopn}

\usepackage{subfigure}
\usepackage{epsfig}
\usepackage{graphicx}

\usepackage{amssymb}

\usepackage{url}
\usepackage[colorlinks,linkcolor=red]{hyperref}

\DeclareMathOperator*{\argmax}{argmax}

\ifCLASSINFOpdf
\else
\fi
\begin{document}
\title{Discriminative Triad Matching \\and Reconstruction for Weakly Referring Expression Grounding}

\author{Mingjie~Sun,
        Jimin~Xiao,~\IEEEmembership{Member,~IEEE,}
        Eng Gee~Lim,~\IEEEmembership{Member,~IEEE,} \\
        Si~Liu,~\IEEEmembership{Member,~IEEE}
        and John Y. Goulermas, ~\IEEEmembership{Senior Member,~IEEE}
\IEEEcompsocitemizethanks{\IEEEcompsocthanksitem M. Sun is with University of Liverpool, UK, and also with School of Advanced Technology, Xi’an Jiaotong-Liverpool University, Suzhou, P.R. China(e-mail:mingjie.sun@liverpool.ac.uk).
\IEEEcompsocthanksitem J. Xiao is with the School of Advanced Technology , Xi'an Jiaotong-Liverpool University, Suzhou, China (Corresponding author:Jimin Xiao, e-mail:jimin.xiao@xjtlu.edu.cn).
\IEEEcompsocthanksitem E. G. Lim is with the School of Advanced Technology, Xi'an Jiaotong-Liverpool University, Suzhou, China (e-mail:enggee.lim@xjtlu.edu.cn).
\IEEEcompsocthanksitem S. Liu is with the Institute of Artificial Intelligence, Beihang University, Beihang University, Beijing, China (e-mail:liusi@buaa.edu.cn).
\IEEEcompsocthanksitem G. Yannis is with the Department of Computer Science, University of Liverpool, Liverpool, UK (e-mail:goulerma@liverpool.ac.uk).
\IEEEcompsocthanksitem The work was supported by National Natural Science Foundation of China under 61972323, and Key Program Special Fund in XJTLU under KSF-T-02, KSF-P-02.
}
}

\markboth{submitted to IEEE TPAMI}%
{Shell \MakeLowercase{\textit{et al.}}: Bare Demo of IEEEtran.cls for Computer Society Journals}

\IEEEtitleabstractindextext{%
\begin{abstract}

In this paper, we are tackling the weakly-supervised referring expression grounding task, for the localization of a referent object in an image according to a query sentence, where the mapping between image regions and queries are not available during the training stage. 
In traditional methods, an object region that best matches the referring expression is picked out, and then the query sentence is reconstructed from the selected region, where the reconstruction difference serves as the loss for back-propagation. The existing methods, however, conduct both the matching and the reconstruction approximately as they ignore the fact that the matching correctness is unknown. 
To overcome this limitation, a discriminative triad is designed here as the basis to the solution, through which a query can be converted into one or multiple discriminative triads in a very scalable way.
Based on the discriminative triad, we further propose the triad-level matching and reconstruction modules which are lightweight yet effective for the weakly-supervised training, making it three times lighter and faster than the previous state-of-the-art methods. 
One important merit of our work is its superior performance despite the simple and neat design. Specifically, the proposed method achieves a new state-of-the-art accuracy when evaluated on RefCOCO (39.21\%), RefCOCO+ (39.18\%) and RefCOCOg (43.24\%) datasets, that is 4.17\%, 4.08\% and 7.8\% higher than the previous one, respectively. The code is available at \url{https://github.com/insomnia94/DTWREG}.
\end{abstract}

}

\maketitle

\IEEEdisplaynontitleabstractindextext
\IEEEpeerreviewmaketitle

\IEEEraisesectionheading{\section{Introduction}
\label{sec:introduction}}
Referring expression grounding (REG) is a fundamental multi-modality task aiming at the recognition and localization of a target object in an image, according to its query sentence (the referring expression). Thus, REG can be utilized throughout many downstream tasks, including the visual question answering (VQA) \cite{7934440,8046084}, visual commonsense reasoning (VCR) \cite{8691415}, visual navigation \cite{4767983}, etc.

Traditional REG methods are trained in a supervised way \cite{yu2018mattnet,yang2019cross}, where the mapping between proposals and queries is available during the training stage. However, drawing the connection between each query sentence and its corresponding proposal is notably time-consuming. Therefore, it has practical significance to handle the REG task in a weakly-supervised setting (WREG), where the aforementioned mapping is no longer available during training.

Generally, existing WREG methods consist of two steps; namely, the sentence-level matching and the reconstruction \cite{liu2019knowledge,liu2019adaptive, guo2020normalized,ge2019exploring}. In the first step, WREG methods roughly assume the sentence-level matching procedures from existing fully-supervised REG methods \cite{yu2018mattnet} in order to calculate the similarity between the entire query and each candidate proposal. Specifically, they either adopt an oversimplified sentence-level matching module where a query is roughly parsed into a subject word and an object word \cite{liu2019knowledge}, making it thus cumbersome to handle complex queries with many descriptive terms, or adopt an overcomplicated matching module with multiple language processing sub-networks attached to analyze complex queries \cite{liu2019adaptive}. Although these supererogatory sub-networks barely hinder fully-supervised training, they become problematic under the weakly-supervised setting, as the correctness of the matching result is not guaranteed and this makes the preceding short back-propagation (BP) path unavailable and the new path longer and noisier. Therefore, the simple model tends to perform better than a complicated one \cite{zhukov2019cross,xiong2019pretrained}. 
The predicament arising here is that a good WREG network requires not only a strong and all-purpose matching module to deal with various queries, and especially so with fragmentary and complex ones, but also a delicate and easy-to-train matching module to facilitate the necessitated weakly-supervised training.

\begin{figure}[t]
	\begin{center}
		\includegraphics[width=1.02\linewidth]{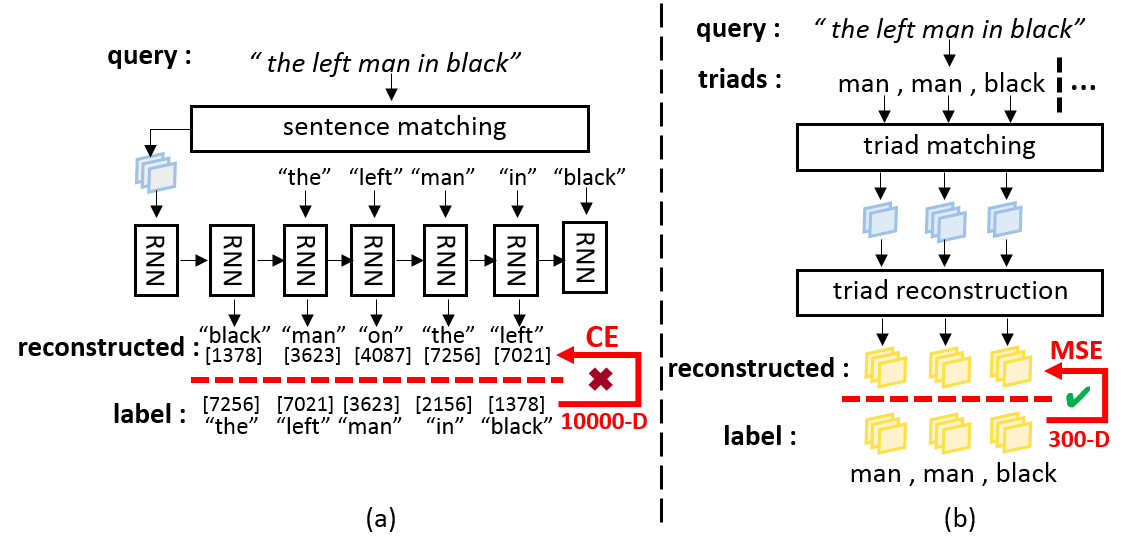}
	\end{center}
	\caption{Illustration of the difference between (a) the traditional WREG method, and (b) the proposed triad-level method. The blue and the yellow parts indicate the proposal features and the linguistic feature, while the red represents the individual loss function. In (a), although the meaning of the reconstructed sentence is the same as the label, the loss is large, as it is calculated word-by-word, making the network hard to converge. In (b) however, the reconstruction is conducted in the triad-level, and a 300-D MSE loss function replaces the 10,000-D CE loss function which dramatically facilitates the weakly-supervised training.}
	\label{intuitive}
\end{figure}

The second step is to build a BP path for weakly-supervised training, where a reconstruction stage is created to rebuild the query information with the difference between the original and reconstructed information being the BP loss. Such difference can be calculated either at a sentence level by rebuilding the entire sentence word-by-word, or based on the extracted key-word features. All existing WREG methods adopt the former one, where a descriptive sentence is predicted from an image region via an RNN-style network, as shown in Fig.\ref{intuitive}.(a). Its accuracy, however, proves hardly satisfactory even for a full-supervised setting \cite{guo2020normalized,ge2019exploring} and makes the BP loss unreliable. Additionally, we have the architectural imbalance from the heavy RNN-style reconstruction network, which is never used in the final inference stage while occupying a large proportion of parameters of the entire network (around 75\% in \cite{liu2019knowledge,liu2019adaptive}).

To address these issues simultaneously, we design a discriminative triad as well as a scalable query parsing strategy where a query sentence can be represented by one or multiple discriminative triads with similar formulations. Specifically, the discriminative triad is a 3-unit set, with the first unit representing the target object, the second unit denoting the reference object against the target object, and the third unit indicating the \emph{discriminative relation} between the target object and the reference object. Note that \emph{discriminative relation} is defined in quite a broad sense, including not only the straightforward relation information extracted from a subject-relation-object query, such as ``the cat on the table'' with the triad \{cat, table, on\}, but also concealed information hidden in a unary query with descriptive terms, such as ``the left man in black''. This query can be parsed to the two triads \{man, man, black\} and \{man, man, left\}; see Fig.\ref{intuitive}.(b). In this way, the discriminative triad can represent various forms of queries to facilitate the matching process.

Contrary to the traditional sentence-level matching modules adopted in the first step of WREG, we propose a triad-level matching model concentrating on the triad-proposal similarity, which proves faster and easier to be well trained under a weakly-supervised setting. Despite its simplicity, complex queries can be handled very effectively, as these can be parsed into multiple discriminative triads where the final bounding box results are selected by exhaustively considering the matching score of each triad. 
Accordingly, rather than reconstructing the entire query sentence word-by-word for the second step, we design a triad-level construction module, where the loss is calculated on the linguistic feature of three triad units serving as direct \emph{shortcuts} between the original query and the reconstructed one. This improves the reliability of the loss function and therefore facilitates the weakly-supervised training; see Fig.\ref{intuitive}.(b).

Ultimately, these triad-level matching and reconstruction modules constitute the discriminative WREG network proposed here that simultaneously provides a light, fast and accurate baseline framework for future WREG works. Our main contributions are summarized as:
\begin{itemize}
	\item
We propose a discriminative triad and a scalable query parsing strategy, where a query sentence (no matter how simple, fragmentary or complex it is) can be converted into one or multiple discriminative triads with the same formulation, with each triad instructing how to recognize a target object from a reference object.
	\item
Based on the discriminative triad, the triad-level matching and reconstruction modules are specifically designed for the WREG task. As the conventional attention-based analysis is avoided, and the word-by-word reconstruction for the entire sentence is replaced by rebuilding three key units in a triad, the proposed method is three times lighter and faster than the previous state-of-the-art (SOTA) method \cite{liu2019knowledge}.
	\item
New SOTA accuracy is obtained on a variety of established REG datasets. Specifically, the proposed network boosts the accuracy by 4.17\%, 4.08\%, and 7.8\% over the best previous SOTA method \cite{liu2019knowledge} when evaluated on RefCOCO, RefCOCO+ and RefCOCOg datasets, respectively.
\end{itemize}

\section{Related work}

\subsection{Supervised REG}
Supervised REG localizes an image region by a query sentence, where the mapping between proposals and queries are available during the training stage. Traditionally, the entire sentence is encoded through a single language embedding network \cite{yu2016modeling}. Considering the variance among different components in a query, Yu \textit{et al.}~\cite{yu2018mattnet} propose an attention mechanism to decompose the query into three linguistic components, describing subject appearance, location and relationship to other objects. Each linguistic component generates a matching score with a candidate proposal, leading to three matching scores for each proposal, and the final result is selected by comprehensively considering all three matching scores. To prevent the attention mechanism from solely focusing on the most dominant features of both modalities, an erasing mechanism is proposed in \cite{liu2019improving}, where the most dominant linguistic or visual information is discarded to drive the model to discover more complementary linguistic-visual correspondences. To better exploit the relationship between the target object and its neighboring objects, Wang \textit{et al.}~\cite{wang2019neighbourhood} propose a graph-based network, in which nodes correspond to object regions and edges represent relations between these objects, so that the object feature representation can be enriched with the additional relation information of its neighborhood.

\subsection{Weakly-supervised REG}
During the training in WREG there is no mapping between image regions and the query sentence. Rohrbach \textit{et al.}~\cite{rohrbach2016grounding} first predict the attention scores between the query and all candidate proposals, with the query reconstructed using the weighted sum over the visual features of all proposals according to their attention scores. The difference between the original query and the reconstructed one is used for training. In \cite{chen2018knowledge}, the location parameters of each candidate proposal are reconstructed and the difference between the original location and the reconstructed one is used for training. Inspired by Mattnet \cite{yu2018mattnet}, where a query is decomposed into three linguistic components, Liu \textit{et al.}~\cite{liu2019adaptive} optimize the network by minimizing the distance between the language feature of each linguistic component and its corresponding visual feature, apart from the sentence-level reconstruction loss. Liu \textit{et al.} \cite{liu2019knowledge} parse each query into a subject-object pair, and the matching process is conducted between the linguistic subject-object pair and visual proposal-proposal pair, while the BP procedure is still based on the sentence-level reconstruction. Differently from these WREG methods, our method avoids the complicated attention-based analysis and RNN-based reconstruction for the entire sentence, and proposes the triad-level matching and reconstruction modules for superior performance.

\section{Methodology}

\begin{figure*}[ht]
	\centering
	\includegraphics[width=0.9 \linewidth]{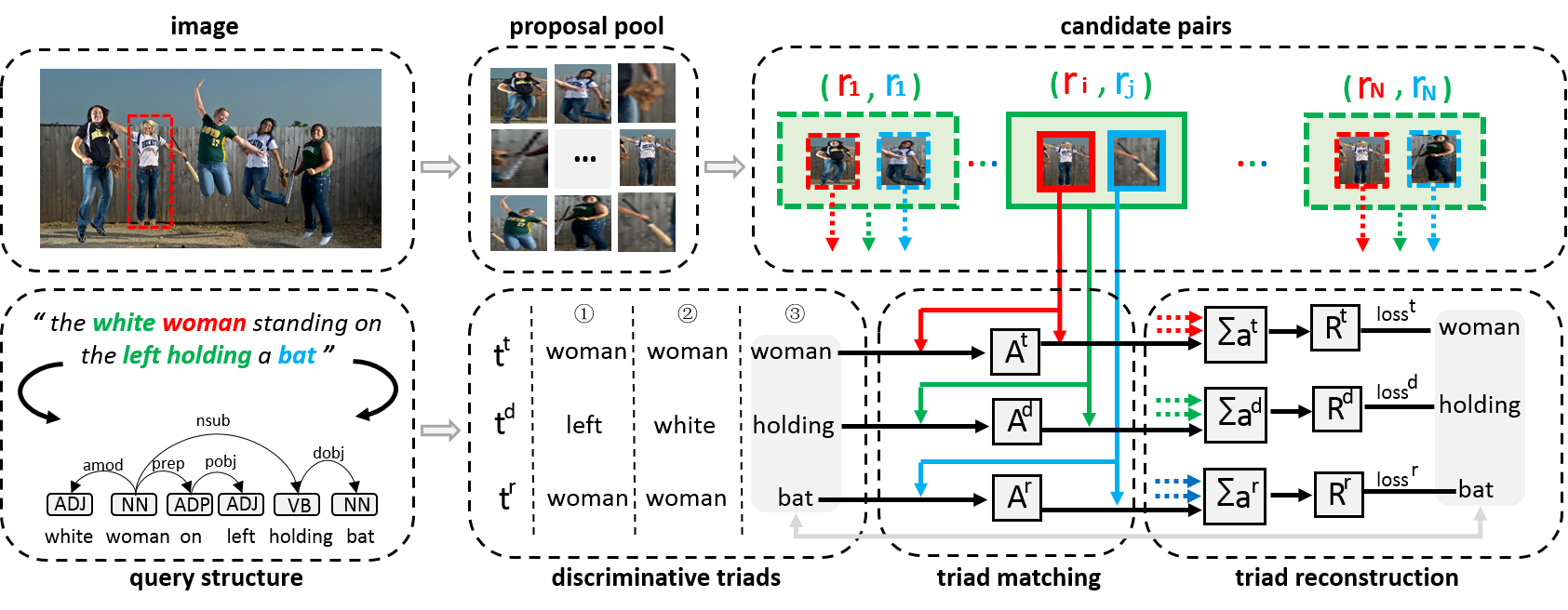}
	\caption{The proposed framework mainly consists of the generation of discriminative triads, the triad matching module and the triad reconstruction module (rightmost three modules in the bottom row). Other modules are also illustrated, including the analysis of the sentence structure (leftmost module in the bottom row) and the organization of proposal pairs (modules in the top row). The parts in solid red, green and blue boxes represent the features of a proposal pair used to match the linguistics feature of the target unit, the discriminative unit and the reference unit in a discriminative triad, respectively. Note that the parts in dashed boxes, whose specific routes are omitted, work in the same way as the parts in solid boxes.}
	\label{framework}
\end{figure*}

WREG can be formulated as a region-level retrieval problem. Given an image $I$, a set of image regions (proposals) $R = \{r_i\}_{i=1}^{N}$ are provided by existing bounding box annotations, or predicted by a region proposal network (RPN) \cite{zitnick2014edge}, where $r$ is the 4-D vector denoting the proposal's top-left and bottom-right corners, and $N$ is the number of proposals in image $I$. The WREG task is to retrieve the target region $r^*$ according to the query sentence $q$ by maximizing the similarity score $S(r_i,q)$, between all candidate proposals $r_i$ and the query $q$, that is
\begin{equation}
 r^*=\argmax\limits_{r\in R}S(r,q).
\end{equation}
In this work, the query sentence $q$ is represented by a set of discriminative triads $T = \{t_k\}_{k=1}^{M}$, where $M$ is the number of such triads. Thus, the final similarity score can be calculated as the sum of the individual scores as
\begin{equation}
S(r,q)=\sum _{k=1}^{M} S(r,t_k).
\end{equation}
Consequently, the target is to find a proper scoring mechanism $S(\cdot,\cdot)$ to correctly distinguish the target region from others. Specifically, as the connection annotation for any region-query pair is unknown during training, a reconstructed query $\hat{t}_k$ is predicted from $r^*$ through a reconstruction module, and the training loss is the difference between $t_k$ and $\hat{t}_k$. The whole network architecture is shown in Fig.\ref{framework}.

\subsection{Discriminative Triad}
\label{sec:triad}
To take full advantage of the discriminative information hiding in queries, we propose to parse and convert a query $q$ into multiple discriminative triads $\{t_k\}_{k=1}^{M}$, each representing a piece of discriminative information to distinguish the target from the distracting objects. A discriminative triad $t_k$ consists of three components, that is, a target unit $t^t_k$, a reference unit $t^r_k$ and a discriminative unit $t^d_k$, indicating the target object, the reference object against the target object, and the discriminative relation between them, respectively. 

As can be observed from Table \ref{triad}, the discriminative unit $t^d_k$ is defined in a broad sense, which can be extracted from not only subject-relation-object style queries in a straightforward way, such as ``the man standing on the table'', but also the unary queries containing hidden discriminative connection information, such as ``black cat''. In addition, some particular tokens are utilized for special referring forms. For instance, the discriminative unit of single-word queries (e.g., ``man'') is set as ``SELF'', and ``UKN'' is used to represent the target unit and reference unit for fragmentary queries (e.g., ``left''). Furthermore, complex queries can be separated into several triads. In this way, discriminative triads can represent a diversity of referring forms, no matter whether they are unary, fragmentary or complex. 

For its implementation, an off-the-shelf NLP processing toolbox (Stanford CoreNLP\cite{chen2014fast}, Spacy\cite{honnibal2017spacy}) is employed to analyze the tree structure of each query, as well as the POS tag and the dependency label \cite{de2006generating} of each word in the query, to generate discriminative triads. In general, triads are generated within two steps. The first step is to find the target unit shared by all triads. Specifically, the rightmost Normal Noun (NN) of the bottom-left Noun Phrase (NP) in the sentence tree is regarded as the target unit. The second step is to generate the reference unit and discriminative unit for each triad. Phrases in different formations are processed in different parsing patterns. For the unary phrase, such as ``the white man'' and ``the left cat'', the first component of the dependency, whose second component is the same as the target unit, except nominal subject (nsubj), prepositional modifiers (prep) and determiner modifiers (det), is viewed as a discriminative unit, with its corresponding reference unit the same as the target unit, indicating a hidden reference object. For the subject-relation-object phrase, such as ``the man holding a cat'', the second component of the dependency, whose first component is the same as the target unit, and the dependency type is nsubj, is viewed as a discriminative unit, with the second component of the dependency, whose first component is the same as that of the current discriminative unit, is viewed as its corresponding reference unit.

\renewcommand\arraystretch{1.5}
\begin{table}[]
	\caption{Common query sentences and their corresponding discriminative triads. \textbf{ID} is the index of the discriminative triad of a query, as a complex query may generate multiple discriminative triads. \textbf{T/R/D-Unit} indicates the target, reference, and discriminative unit, respectively.}
	\begin{center}
		\begin{tabular}{|p{80pt}<{\centering} |p{7pt}<{\centering} |p{30pt}<{\centering} p{30pt}<{\centering} p{30pt}<{\centering}|}
			\hline
			\textbf{Query} & \textbf{ID} & \textbf{T-Unit} & \textbf{R-Unit}  & \textbf{D-Unit}\\
			\hline
			``man''               & 1   & man     & man    & SELF  \\
			\hline
			``left''              & 1   & UKN     & UKN    & left  \\
			\hline
			``left man''       & 1  & man     & man    & left  \\
			\hline
			``black cat''     & 1   & cat     & cat    & black  \\
			\hline
			``cat on a table''    & 1  & cat     & table    & on  \\
			\hline
			``man holding a cat''     & 1 &  man    & cat    & holding  \\
			\hline
			\multirow{5}*{\shortstack{``the left man in black \\ holding a red cat and \\ standing on a table''}}       & 1  & man     & man    & left  \\
			~       & 2  & man     & man    & black  \\
			~       & 3  & man     & table    & on  \\
			~       & 4  & man     & cat    & holding  \\
			~       & 5  & cat     & cat    & red  \\
			\hline
		\end{tabular}
	\end{center}
	\label{triad}
\end{table}

\subsection{Triad-level Matching}
The first stage of the triad-level matching is to encode the image $I$ and query $q$. To encode an image $I$, given a set of candidate proposals $\{r_i\}_{i=1}^{N}$, $I$ is first passed into a convolutional network (e.g., ResNet \cite{he2016deep}), with its last convolutional layer output as the visual feature $v_I$ of image $I$. The visual feature $f^v_i$ of proposal $r_i$ is cropped within $v_I$ through a pooling mechanism (e.g., ROI pooling \cite{ren2015faster}). Then, following previous works \cite{yu2016modeling}, the spatial feature of each proposal $r_i$ is encoded by a 5-D vector $f^s_i=[\frac{x_{tl}}{W}, \frac{y_{tl}}{H}, \frac{x_{br}}{W}, \frac{y_{br}}{H}, \frac{w\cdot h}{W\cdot H}]$, where $[\frac{x_{tl}}{W}, \frac{y_{tl}}{H}]$ and $[\frac{x_{br}}{W}, \frac{y_{br}}{H}]$ refer to the top-left and bottom right corners of $r_i$, respectively. $w$ and $h$ indicate the width and height of $r_i$, while $W$ and $H$ denote the width and height of $I$. In this way, $f^s_i$ contains the absolute position information of $r_i$, as well as its relative area size to $I$. To encode a query sentence $q$ with its discriminative triads $\{t_k\}_{k=1}^{M}$  = $\{ (t^t_k,  t^r_k, t^d_k) \}_{k=1}^{M}$, a pre-trained word vector mechanism (e.g., Word2vec\cite{mikolov2013efficient}, Glove\cite{pennington2014glove}) is employed to extract the linguistic embedding $e^t_k$, $e^r_k$, $e^d_k$ with the size of $D_l$, corresponding to $t^t_k$, $t^r_k$, $t^d_k$, respectively.

The next stage of the triad-matching is to predict the similarity score between $q$ and each candidate proposal $r_i$ in image $I$ through the matching module. As $q$ is parsed and converted into a set of discriminative triads $\{t_k\}_{k=1}^{M}$, the matching module is designed to focus on the single discriminative triad $t_k$, in order to facilitate the training in the weakly-supervised setting. Furthermore, since an individual discriminative triad $t_k$ comprises target unit $t^t_k$, reference unit $t^r_k$ and discriminative unit $t^d_k$, the purpose of the matching module turns to find a suitable proposal pair $(r_i,r_j)$ of $I$, matching $t^t_k$, $t^r_k$ and $t^d_k$ based on three similarity scores predicted by the matching module. 

Specifically, to predict these scores, three unit-level attention-based matching modules are designed, including the target attention module $A^t$, the reference attention module $A^r$ and the discriminative attention module $A^d$. The target attention module $A^t$ outputs a score $a^t_{k,i}=A^t(f^v_{i},e^t_k;\Theta_t)$ representing how likely the first proposal $r_i$ in $(r_i,r_j)$ matches the target unit $t^t_k$, based on
\begin{equation}
a^t_{k,i} = W^t_2\phi_{ReLU}\Big(W^t_1\phi_{ReLU} (f^v_{i}\oplus e^t_k)   +b^t_1\Big)+b^t_2,  
\label{ep:attention_t}
\end{equation}
where $W^t_1$, $W^t_2$, $b^t_1$, $b^t_2$ are parameters of the target attention module weight $\Theta_t$, $\phi_{ReLU}(\cdot)$ indicates the ReLU activation function, and $\oplus$ is a concatenation process. 

Similarly, the reference attention module $A^r$ predicts a score $a^r_{k,j}=A^r(f^v_{j},e^r_k;\Theta_r)$ representing how likely the second proposal $r_j$ matches the reference unit $t^r_k$, as
\begin{equation}
a^r_{k,j} = W^r_2\phi_{ReLU}\Big(W^r_1\phi_{ReLU} (f^v_{j}\oplus e^r_k)   +b^r_1\Big)+b^r_2.
\label{ep:attention_r}  
\end{equation}
Finally, the discriminative attention module $A^d(f^p_{i,j},e^d_k;\Theta_d)$ predicts the attention score between the entire proposal pair $(r_i,r_j)$ and the discriminative unit $t^d_k$, according to
\begin{equation}
a^d_{k,i,j} = W^d_2\phi_{ReLU}\Big(W^d_1\phi_{ReLU} (f^{p}_{i,j}\oplus e^d_k)   +b^d_1\Big)+b^d_2,
\label{ep:attention_d}  
\end{equation}
where $f^{p}_{i,j}$ is a concatenation of \{$f^v_{i}$, $f^s_{i}$, $f^v_{j}$, $f^s_{j}$\} containing all information of both $r_i$ and $r_j$ to better extract their interrelated connection information.

\subsection{Triad-level Reconstruction}
\label{sec:reconstruct}
Based on the matching scores, a triad-level reconstruction module is designed to rebuild the discriminative triad. The difference between the linguistic feature of the reconstructed triad and the original one is used to optimize the entire network, including both the triad-level matching and the reconstruction modules.

As a single discriminative triad $t_k$ includes three triad units (i.e., $t^t_k$, $t^r_k$ and $t^d_k$), the specific reconstruction target turns to rebuild the linguistic feature of each unit, by three corresponding different modules, $R^t$, $R^r$ and $R^d$. Two different methods are designed to generate the input of each unit-level reconstruction module, including the soft and hard methods.

In the \emph{soft method}, the input is the weighted sum over the features of all candidate proposals or proposal pairs, aggregated according to corresponding unit-level attention scores:
\begin{equation}
\begin{split}
\hat{f}^t_k&=\sum_{i=1}^{N}a^t_{k,i}f^v_{i} \\
\hat{f}^r_k&=\sum_{j=1}^{N}a^r_{k,j}f^v_{j} \\
\hat{f}^d_k&=\sum_{i=1}^{N}\sum_{j=1}^{N}a^d_{k,i,j}f^p_{i,j},
\end{split}
\label{agg}
\end{equation}
where $\hat{f}^t_k$, $\hat{f}^r_k$, and $\hat{f}^d_k$ represent the input of the unit-level reconstruction modules, $R^t$, $R^r$ and $R^d$, respectively. 

In the \emph{hard method}, the input is the feature of the $(r_i,r_j)$ pair with the highest attention score. Since $argmax$ is not differentiable, it is replaced with the differentiable $softmax$ operator through the Gumbel-Softmax \cite{jang2016categorical}
\begin{equation}
\hat{a}^c = softmax(a^c/\tau),
\end{equation}
where $c \in \{  t,r,d \}$ is the category of a triad unit, and $a^c$ represents the corresponding attention values calculated by Eqs.~(\ref{ep:attention_t}, \ref{ep:attention_r}, \ref{ep:attention_d}). $\tau$ is a temperature parameter allowing $softmax$ to approache $argmax$ as $\tau \rightarrow  0$. $\hat{f}^t_k$, $\hat{f}^r_k$, and $\hat{f}^d_k$ are re-evaluated with the new attention value $\hat{a}^c$  using Eq.(\ref{agg}).

After the input is determined by either the soft method or the hard method, contrary to existing works rebuilding the entire query with a heavy RNN-style model, the proposed reconstruction module is designed to reconstruct the linguistic feature of each individual triad unit (i.e., $e^t_k$, $e^r_k$ and $e^d_k$) via
\begin{equation}
	\hat{e}^c_k=R^c(\hat{f}_k^c),
\end{equation}
where $c \in \{  t,r,d \}$ is the category of a triad unit, $\hat{e}^c$ represents the reconstructed linguistic feature of its corresponding triad unit, and $R^c$ is a lightweight unit-level reconstruction module, which is easier to be trained in the weakly-supervised setting. The triad-based reconstruction module is a simple network, consisting of two fully connected layers with the output size as $D_l$.

\subsection{Training and Inference}
\label{inference}

The entire discriminative network, including the matching module and reconstruction module, is optimized according to the sum of $L_2$ distances between all original triad units (i.e., $e^t_k$, $e^r_k$ ,$e^d_k$) and corresponding predicted ones (i.e., $\hat{e}^t_k$, $\hat{e}^r_k$, $\hat{e}^d_k$). The loss function $\mathbb{L}$ is given by
\begin{equation}
	\mathbb{L} = \sum_{c \in \{  t,r,d \}} \left \| \hat{e}^c_k - e^c_k \right \|^2_2.
\label{loss}
\end{equation}

During the inference stage, for a certain proposal pair $(r_i,r_j)$, its triad-level attention score $\bar{a}_{k,i,j}$, representing how likely $(r_i,r_j)$ matches the entire discriminative triad $t_k$, is defined as the weighted sum of three unit-level attentions scores
\begin{equation}
	\bar{a}_{k,i,j} = \alpha a^t_{k,i} + \beta a^r_{k,j} + \gamma a^d_{k,i,j},
	\label{inference_eval}
\end{equation}
where $\alpha$, $\beta$ and $\gamma$ are the hyper-parameters to control the weight of different attention scores. Then, for a certain proposal $r_i$, its triad-level attention score $\bar{a}_{k,i}$ is defined as the highest score of the proposal pairs, where $r_i$ is the first element, as
\begin{equation}
	\bar{a}_{k,i}= \max_{r_j \in R} \; \bar{a}_{k,i,j}.
\end{equation}
Ultimately, for a certain proposal $r_i$, its sentence-level attention score $\bar{a}_{i}$ is defined as the sum of triad-level attention scores over all discriminative triads through
\begin{equation}
\bar{a}_{i}= \sum_{k=1}^{M} \; \bar{a}_{k,i},
\end{equation}
and the proposal $r^*$ with the highest sentence-level attention score is selected as the predicted result: 
\begin{equation}
	i^* = \argmax\limits_{r \in R} (\bar{a}_{i}).
\end{equation}

\section{Experiments}

\subsection{Implementation Details}

The candidate proposals used in this work can be either provided from the bounding box ground-truth in MS-COCO dataset \cite{lin2014microsoft}, or generated by the detection network, Faster RCNN \cite{ren2015faster}, pre-trained on MS-COCO. Stanford CoreNLP \cite{chen2014fast} is adopted to parse queries. In the feature encoding stage, the image-level visual feature $v_I$ is extracted through the last convolutional layer of Resnet101 \cite{he2016deep}. The individual visual feature of each candidate proposal is cropped within $v_I$ using ROI pooling \cite{ren2015faster}. Glove word vector \cite{pennington2014glove} is adopted to extract the linguistic feature of the discriminative triad. The discriminative network is trained using one Titan X GPU for around 20 hours adopting the ADAM optimizer \cite{kingma2014adam}. The learning rate is set as 1.3e-5, and the iteration number is set as 150,000 (3 epochs) for all training sets. In the hard reconstruction method, $\tau$ is set to initiate the hard reconstruction method. During the inference time, we set ($\alpha$,$\beta$,$\gamma$)= (2,1,1) in Eq.(\ref{inference_eval}) as the target attention score directly contributes to localizing the target object, while the reference or discriminative attention score only serves as auxiliary information.

\begin{table*}[]
	\caption{Effectiveness comparison with other WREG methods. The method with the highest score is in bold. ``det'' indicates that the candidate proposals generated from Faster RCNN \cite{ren2015faster}. In other cases, the candidate proposals are provided by the bounding box ground truth.}
	\begin{center}
		\begin{tabular}{|p{70pt}<{\centering}|p{40pt}<{\centering}|p{30pt}<{\centering}  p{30pt}<{\centering}  p{30pt}<{\centering}  |p{30pt}<{\centering}  p{30pt}<{\centering}  p{30pt}<{\centering}  |p{50pt}<{\centering} |}
			\hline
			\multirow{2}{*}{Method} & \multirow{2}{*}{Setting} & \multicolumn{3}{c|}{RefCOCO} & \multicolumn{3}{c|}{RefCOCO+} & RefCOCOg \\ \cline{3-9} 
			&                           & val    & testA    & testB    & val     & testA    & testB    & val      \\ \hline \hline
			VC\cite{rohrbach2016grounding} \tiny{CVPR18}         & -                         & -         & 17.34    & 20.98    & -         & 23.24    & 24.91    & 33.79    \\
			VC\cite{rohrbach2016grounding} \tiny{CVPR18}         & w/o reg                   & -         & 13.59    & 21.65    & -         & 18.79    & 24.14    & 25.14    \\
			VC\cite{rohrbach2016grounding} \tiny{CVPR18}         & w/o a                     & -         & 33.29    & 30.13    & -         & 34.60    & 31.58    & 30.26    \\
			VC (det)         &  -             & -     & 20.91    & 21.77    & -     & 25.79    & 25.54    & 33.66    \\ \hline
			ARN\cite{liu2019adaptive} \tiny{ICCV19}        & -                         & 34.26     & 36.01    & 33.07    & 34.53     & 36.01    & 33.75    & 34.66    \\
			ARN\cite{liu2019adaptive} \tiny{ICCV19}        & w/o $L_{lan}$             & 33.07     & 36.43    & 29.09    & 33.53     & 36.40    & 29.23    & 33.19    \\
			ARN\cite{liu2019adaptive} \tiny{ICCV19}        & w/o $L_{att}$             & 33.60     & 35.65    & 31.48    & 34.40     & 35.54    & 32.60    & 34.50    \\ 
			ARN (det)         &  -             & 32.17     & 35.25    & 30.28    & 32.78     & 34.35    & 32.13    & 33.09    \\ \hline
			KPRN\cite{liu2019knowledge} \tiny{ACMMM19}      & -                         & 35.04     & 34.74    & 36.53    & 35.10     & 32.75    & 36.76    & 35.44    \\
			KPRN\cite{liu2019knowledge} \tiny{ACMMM19}      & attr                      & 34.93     & 33.76    & 36.98    & 35.31     & 33.46    & 37.27    & 38.37    \\
			KPRN\cite{liu2019knowledge} \tiny{ACMMM19}      & soft                      & 34.43     & 33.82    & 35.45    & 35.96     & 35.24    & 36.96    & 33.56    \\  \hline	
			Proposed                     & -                         & \textbf{39.21}     & \textbf{41.14}    & \textbf{37.72}    & \textbf{39.18}     & \textbf{40.01}    & \textbf{38.08}    & \textbf{43.24}    \\ 
			Proposed (det)         &  -             & 38.35     & 39.51    & 37.01    & 38.91     & 39.91    & 37.09    & 42.54    \\ \hline
		\end{tabular}
		\label{comparisonCOCO}
	\end{center}
\end{table*}

\begin{figure*}[ht]
	\centering
	\includegraphics[width=0.9 \linewidth]{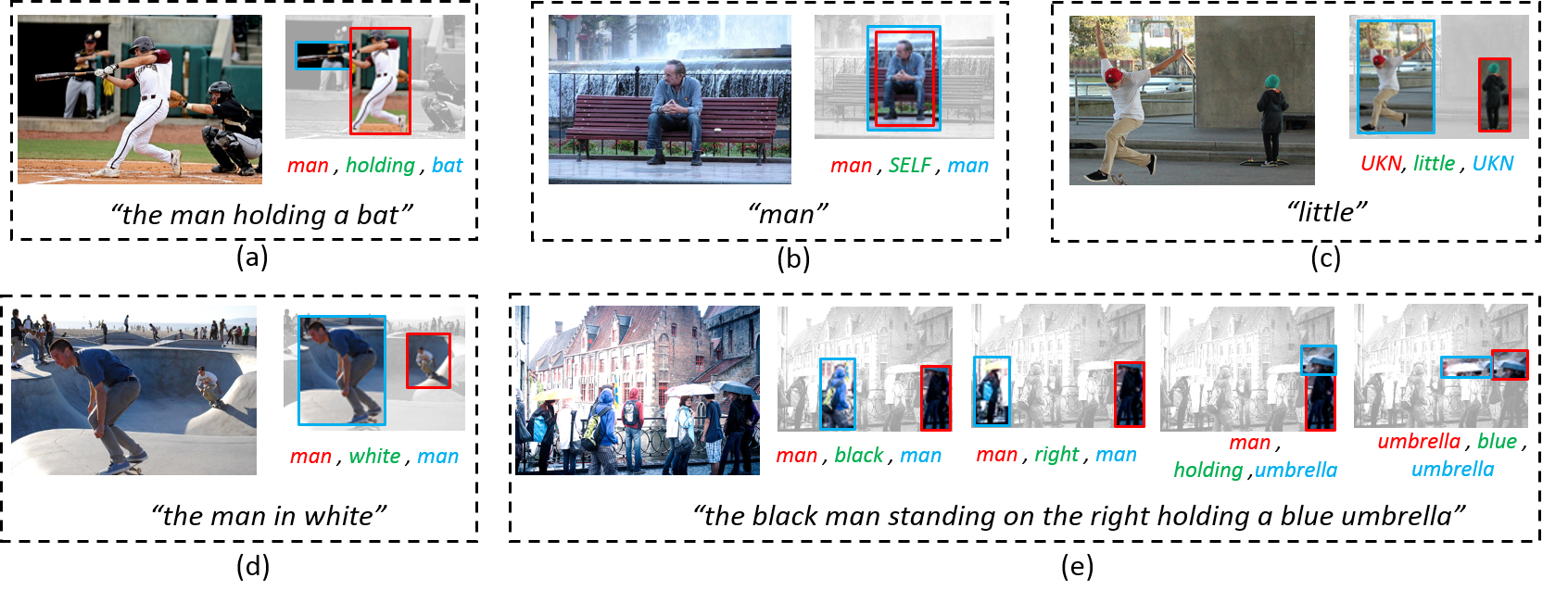}
	\caption{Visualization results on MSCOCO datasets, where the red, green, blue parts represent the target unit, discriminative unit, reference unit and their corresponding proposals, respectively.}
	\label{qualitative}
\end{figure*}

\subsection{Datasets and Metric}
The proposed method is evaluated on four common datasets including RefCOCO \cite{yu2016modeling}, RefCOCO+ \cite{yu2016modeling}, RefCOCOg \cite{mao2016generation} and RefCLEF \cite{kazemzadeh2014referitgame}. The first three datasets are collected from MS-COCO \cite{lin2014microsoft}. The main difference between RefCOCO+ and RefCOCO is that the former one forbids the absolute location words when generating the query sentences, requiring more attention to the appearance discrimination. RefCOCOg contains longer expressions for both appearance and locations. Referring sentences in RefCLEF are annotated casually, so it is normally used as an auxiliary one for validation. The metric used to evaluate the accuracy of the proposed method is similar to the object detection task, that is the Intersection-over-union (IoU) between the predicted bounding box and the ground-truth one is calculated and the one with IoU higher than 0.5 is treated as correct. 

\subsection{Comparison with State-of-the-Art}
In this section, to demonstrate the effectiveness and efficiency of the proposed method, and it is compared in terms of accuracy, parameter cardinality and running speed with other SOTA WREG methods, including KPRN \cite{liu2019knowledge}, ARN \cite{liu2019adaptive} and VC \cite{rohrbach2016grounding}. 

\begin{figure}
	\centering
	\includegraphics[width=1 \linewidth]{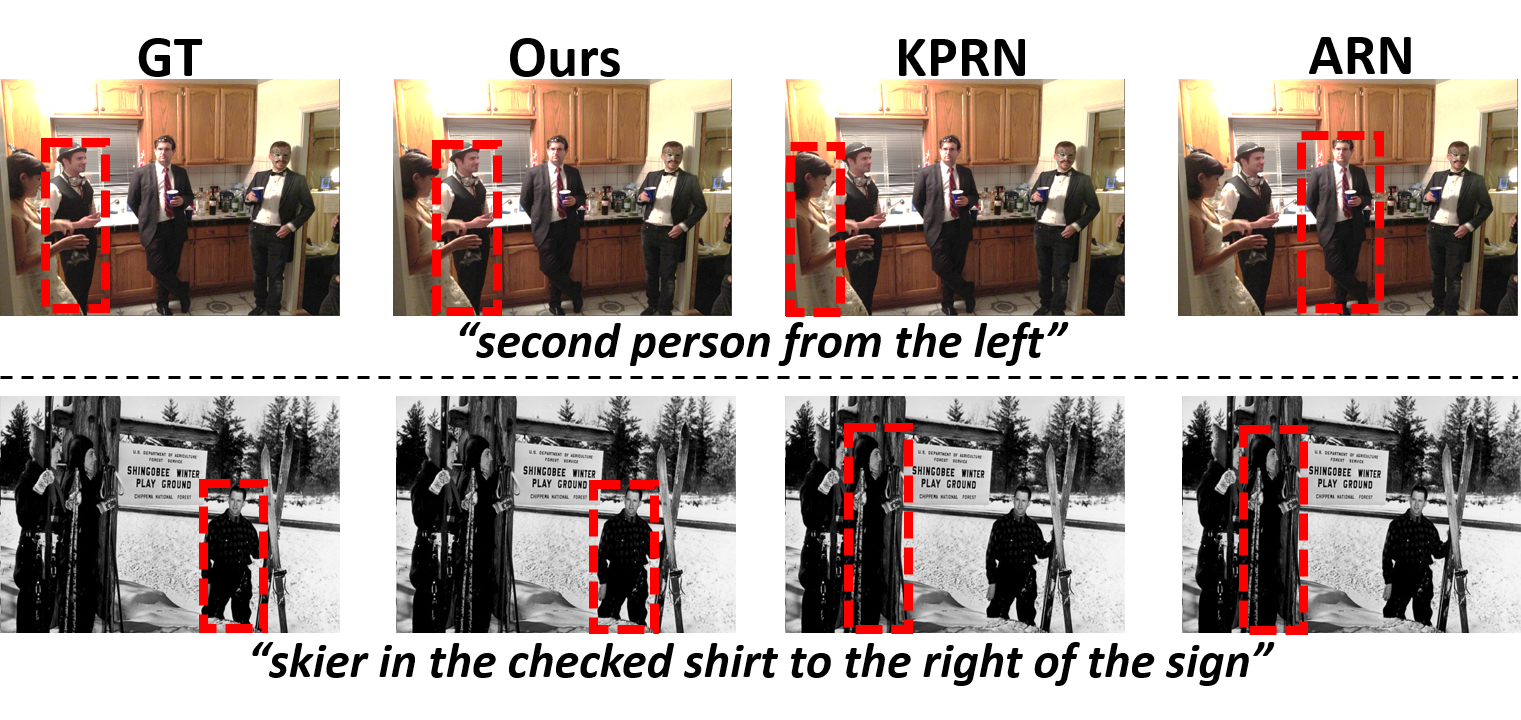}
	\caption{The qualitative comparison with other WREG methods.The red box indicates the bounding box result of the target.}
	\label{comparison}
\end{figure}

In terms of the effectiveness, the accuracy comparison is first conducted between the proposed method and other SOTA WREG methods on three standard datasets, as shown in Table \ref{comparisonCOCO} and Fig.\ref{comparison}, including RefCOCO, RefCOCO+ and RefCOCOg datasets. When the candidate proposals are provided from the bounding box ground truth, the proposed method outperforms previous WREG methods by a large margin on all these three datasets. Specifically, the proposed method boosts the accuracy by 4.17\%, 4.08\% and 7.8\%, against the previous SOTA method, KPRN \cite{liu2019knowledge}, on the valuation set of RefCOCO, RefCOCO+ and RefCOCOg dataset, respectively. Note that the reason for the higher accuracy gain on RefCOCOg is that queries in that dataset are much longer, where more trivial words are removed through triad-based parsing. Furthermore, the proposed method achieves a larger accuracy gain over KPRN \cite{liu2019knowledge} on the Test A set of RefCOCO and RefCOCO+ datasets, which is 6.4\% and 7.26\%, respectively. However, the accuracy gain turns smaller on the Test B set of RefCOCO and RefCOCO datasets. We believe the reason is that, compared with the human targets in the Test A set, the general object targets in the Test B set are hard to describe clearly and unambiguously, making it difficult to localize these targets. The rise of accuracy is also seen when the candidate proposals are generated by a detection network Faster RCNN \cite{ren2015faster}. 

Apart from the accuracy gain, it can also be observed from Table \ref{comparisonCOCO} that the proposed method is more stable and robust, compared with other methods. In other words, differently from previous methods relying on different settings to achieve the best performance on different datasets, the proposed method performs well on all datasets under a single and fixed setting. For instance, KPRN \cite{liu2019knowledge} can achieve its highest accuracy on RefCCOCO dataset under its standard-setting, while for RefCOCO+ dataset, it needs to switch to the \emph{soft} setting to achieve its highest accuracy, and for RefCOCOg, the \emph{attr} setting is required for the best performance. The proposed method, nevertheless, does not need any additional setting, demonstrating its robustness to all datasets. The proposed method is also evaluated on the RefCLEF dataset to further demonstrate its effectiveness. Even with many ambiguous and mistakenly annotated referring expressions in the RefCLEF dataset, the proposed method achieves the accuracy of 34.03\%, which is higher than KPRN by 0.17\%.

In terms of efficiency, the parameter number of the proposed method (7 million) is approximately 1/3 of ARN \cite{liu2019adaptive} and KPRN \cite{liu2019knowledge} (24 and 20 million). In addition, the proposed methods run three times faster than ARN \cite{liu2019adaptive} and KPRN \cite{liu2019knowledge} (0.04 and 0.03 seconds for each picture), demonstrating the efficiency of the proposed system. The major reason is that the heavy RNN-style model is replaced by our proposed model, which is lightweight yet efficient.

Some visualization results of the proposed method are also shown in Fig.\ref{qualitative}, where candidate proposals are provided from bounding box labels. Each diagram, from (a) to (e), illustrates the situation when correspondingly encountering a normal, single-word, fragmentary, reference-hidden or complex query. This demonstrates that the proposed method performs well in a diversity of query formulations.

\subsection{Ablation Studies}
All ablation studies are conducted through cross-evaluation on the training set of RefCOCO, RefCOCO+ or RefCOCOg dataset. Specifically, 3/4 data of the original training set are used as the new training split and the remaining 1/4 data are used as the new valuation split. 

\subsubsection{Contribution of Each Unit-level Loss}
\begin{table}[]
	\caption{Ablation studies through cross-evaluation.}
	\begin{center}
		\begin{tabular}{|c|c|c|c|}
			\hline
			Setting      & RefCOCO & RefCOCO+ & RefCOCOg \\ \hline \hline
			w/o $L^t$    & 29.5        & 27.1       & 27.0          \\
			w/o $L^d$    & 38.1        & 36.1        & 35.5         \\
			w/o $L^r$    & 37.5        & 35.7        & 37.1       \\ \hline
			w/o Recon       & 35.6        & 37.5        & 35.9          \\
			Soft      & 42.1        & 39.2        & 38.2          \\ \hline
			Single       & 39.6        & 38.1        & 36.1          \\ \hline
			Ours         & \textbf{43.2}                 & \textbf{40.1}                 & \textbf{39.8}         \\       
			\hline
		\end{tabular}
		\label{ablation}
	\end{center}
\end{table}
The first ablation study is to explore the contribution of each unit-level loss component in Eq.(\ref{loss}). Specifically, the loss component for the reconstructed target unit, discriminative unit, reference unit is disabled individually during the training stage, generating the setting w/o $L^t$, w/o $L^d$ and w/o $L^r$ in Table \ref{ablation}, respectively. As can be observed from Table \ref{ablation}, the accuracy declines considerably if any unit-level loss component is disabled, which demonstrates their joint and individual contribution to the final performance. Furthermore, the largest decline is seen in the w/o $L^t$ setting. We believe the reason is that, the target unit directly contributes to localize the target object, while the discriminative and reference units only serve as auxiliary information. Based on the aforementioned results, all unit-level loss components are adopted in the proposed method.

\subsubsection{Reconstruction Settings}
The second ablation study is to explore the best setting for the triad-level reconstruction module. In the first setting, the reconstruction module is disabled, with the feature distance between each $(r_i,r_j)$ pair's feature (i.e., $f^v_i$, $f^v_j$, $f^p_{i,j}$), and the corresponding triad unit's linguistic feature (i.e., $e^t_k$, $e^r_k$, $e^d_k$) served as the loss, generating the \emph{w/o Recon} in Table \ref{ablation}. The second and third settings are the soft and hard methods described in section \ref{sec:reconstruct}, generating \emph{Soft} and \emph{Ours} in Table \ref{ablation}. As can be observed in Table \ref{ablation}, the none-reconstruction setting \emph{w/o Recon} performs worst, demonstrating the significance of the triad-level reconstruction module. In addition, we believe the reason for its bad performance is that $f^v_i$, $f^v_j$, $f^p_{i,j}$ and $e^t_k$, $e^r_k$, $e^d_k$ are in two different feature domains, and it is arduous to calculate a meaningful distance before feature alignment. For the comparison of other two settings, as shown in Table \ref{ablation}, \emph{Ours} achieves a relatively higher accuracy against \emph{Soft}. The reason is that the Gumbel-Softmax can focus more on the visual feature with the highest attention value and less on other distracting features, facilitating thus the training of the reconstruction modules.

\subsubsection{Inference Methods}
The third ablation study is to explore the influence of different inference methods. Apart from the standard one described in section \ref{inference} (denoted as \emph{Ours} in Table \ref{ablation}), another baseline setting, where a query is represented by a single discriminative triad, randomly selected among all candidate triads, is also evaluated for better analysis, marked as \emph{Single} in Table \ref{ablation}. As can be observed from Table \ref{ablation}, the bad performance of \emph{Single} demonstrates the significance to take the full advantage of all pieces of discriminative information in a query, especially for some complex queries.

\section{Conclusions}
In this paper, we have proposed a discriminative triad representation, through which a query sentence can be parsed into one or multiple discriminative triads in a very scalable way. Also, based on the discriminative triad, the triad-level matching and reconstruction modules are especially designed for lightweight yet very effective weakly-supervised training. We expect this method can provide a fast and accurate WREG baseline method which can be easily extended and adapted in future works. 


\bibliographystyle{IEEEtran}

\bibliography{IEEEabrv,bare_jrnl}

\end{document}